\documentclass[11pt,a4paper]{article}
\usepackage[hyperref]{acl2019}
\usepackage{times}
\usepackage{latexsym}

\usepackage{url}

\aclfinalcopy 
\def\aclpaperid{69} 

\newcommand\blfootnote[1]{%
  \begingroup
  \renewcommand\thefootnote{}\footnote{#1}%
  \addtocounter{footnote}{-1}%
  \endgroup
}

\usepackage{xcolor}
\usepackage{tabularx}
\usepackage{float}
\restylefloat{table}
\usepackage{url}
\usepackage{algorithm}

\usepackage{mathtools}
\usepackage{varwidth}
\usepackage{algpseudocode}
\usepackage{booktabs}
\usepackage{threeparttable}
\usepackage{enumitem}
\usepackage{comment}
\usepackage{dsfont}

\usepackage{bm,amsmath,amssymb}
\usepackage{xspace}
\usepackage{graphicx}
\usepackage{color}

\usepackage{colortbl}
\definecolor{light_gray}{RGB}{170,170,170}

\usepackage[utf8]{inputenc} 
\usepackage[T1]{fontenc}    
\usepackage{url}            
\usepackage{booktabs}       
\usepackage{amsfonts}       
\usepackage{nicefrac}       
\usepackage{microtype}      
\usepackage{longtable,lscape}
\usepackage{array,multirow}
\usepackage[normalem]{ulem}

\usepackage{subcaption}

\usepackage{wrapfig}

\newcommand{\hide}[1]{}

\usepackage{enumitem}

\usepackage{bm}

\newcommand{\mname}{the augmentation model\xspace}
\newcommand{\set}[1]{\ensuremath \mathcal{#1}}
\newcommand{\reals}{\ensuremath \mathbb{R}}

\title{Clinical Concept Extraction for Document-Level Coding}

 \author{Sarah Wiegreffe$^1$, Edward Choi$^{1*}$, Sherry Yan$^2$, Jimeng Sun$^1$, Jacob Eisenstein$^1$ \\
 $^1$Georgia Institute of Technology \\
 $^2$Sutter Health\\
 $^*$ Current Affiliation: Google Inc\\
     \tt saw@gatech.edu, \tt edwardchoi@google.com, \tt yansx@sutterhealth.org, \\\tt jsun@cc.gatech.edu, \tt jacobe@gatech.edu \\
   }

\date{}

\begin{document}
\maketitle
\begin{abstract}


The text of clinical notes can be a valuable source of patient information and clinical assessments.
Historically, the primary approach for exploiting clinical notes has been information extraction: linking spans of text to concepts in a detailed domain ontology.
However, recent work has demonstrated the potential of supervised machine learning to extract document-level codes directly from the raw text of clinical notes.
We propose to bridge the gap between the two approaches with two novel syntheses: (1) treating extracted concepts as \emph{features}, which are used to supplement or replace the text of the note; (2) treating extracted concepts as \emph{labels}, which are used to learn a better representation of the text. 
Unfortunately, the resulting concepts do not yield performance gains on the document-level clinical coding task.
We explore possible explanations and future research directions.
\end{abstract}

\section{Introduction}
\label{sec:intro}
Clinical decision support from raw-text notes taken by clinicians about patients has proven to be a valuable alternative to state-of-the-art models built from structured EHRs. Clinical notes contain valuable information that the structured part of the EHR does not provide, and do not rely on expensive and time-consuming human annotation~\cite{torres2017icd, clinicalcoding}. Impressive advances using deep learning have allowed for modeling on the raw text alone~\cite{mullenbach2018explainable, rios2018emr, baumel2018multilabel}. However, there exist some shortcomings to these approaches: clinical text is noisy, and often contains heavy amounts of abbreviations and acronyms, a challenge for machine reading \cite{nguyen2016text}. Additionally, rare words replaced with "UNK" tokens for better generalization may be crucial for predicting rare labels.

Clinical concept extraction tools abstract over the noise inherent in surface representations of clinical text by linking raw text to standardized concepts in clinical ontologies. The Apache clinical Text Analysis Knowledge Extraction System (cTAKES, \citealp{ctakes}) is the most widely-used such tool, with over 1000 citations. Based on rules and non-neural machine learning methods and engineered for almost a decade, cTAKES provides an easily-obtainable source of human-encoded domain knowledge, although it cannot leverage deep learning to make document-level predictions.



Our goal in this paper is to maximize the predictive power of clinical notes by bridging the gap between information extraction and deep learning models. We address the following research questions: how can we best leverage tools such as cTAKES on clinical text? Can we show the value of these tools in linking unstructured data to structured codes in an existing ontology for downstream prediction? 

We explore two novel hybrids of these methods: data augmentation (augmenting text with extracted concepts) and multi-task learning (learning to predict the output of cTAKES). 
Unfortunately, in neither case does cTAKES improve downstream performance on the document-level clinical coding task. 
We probe this negative result through an extensive series of ablations, and suggest possible explanations, such as the lack of word variation captured through concept assignment.



\section{Related Work}
\label{sec:related}
\paragraph{Clinical Ontologies}

\begin{figure}
\centering
\includegraphics[width=\linewidth]{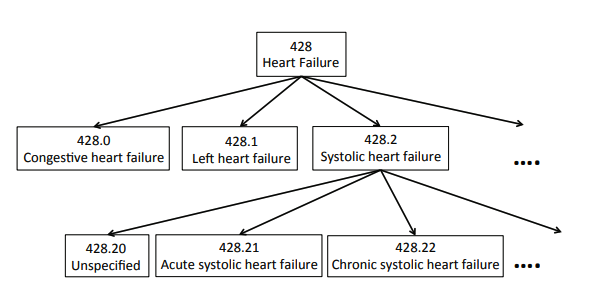}
\caption{A subtree of the ICD ontology \cite[figure from][]{singh2014leveraging}.}
\label{fig:icd}
\end{figure}

Clinical concept ontologies facilitate the maintenance of EHR systems with standardized and comprehensive code sets, allowing consistency across healthcare institutions and practitioners. The Unified Medical Language System (UMLS) \cite{UMLS} maintains a standardized vocabulary of clinical concepts, each of which is assigned a concept unique identifier (CUI). The Systematized Nomenclature of Medicine- Clinical Terms (SNOMED-CT) \cite{SNOMED} and the International Classification of Diseases (ICD) \citep*{ICD9} build off of the UMLS and provide structure by linking concepts based on their relationships. The SNOMED ontology has over 340,000 active concepts, ranging from fine-grained ("Adenylosuccinate lyase deficiency") to extremely general ("patient").
The ICD ontology is narrower in scope, with around 13,000 diagnosis and procedure codes used for insurance billing. Unlike SNOMED, which has an unconstrained graph structure, ICD9 is organized into a top-down hierarchy of specificity (see \autoref{fig:icd}).

\paragraph{Clinical Information Extraction Tools}
There are several tools for extracting structured information from clinical text. Popular types of information extraction include \emph{named-entity recognition}, identifying words or phrases in the text which align with clinical concepts, and \emph{ontology mapping}, labelling the identified words and phrases with their respective clinical codes from an existing ontology.\footnote{Ontology mapping also serves as a form of text normalization.} Of the tools which perform both of these tasks, the open-source Apache cTAKES is used in over 50\% of recent work \cite{wang2017clinical},
outpacing competitors such as MetaMap \cite{metamap} and MedLEE \cite{medlee}.

\begin{figure*}
\centering
\includegraphics[width=\linewidth]{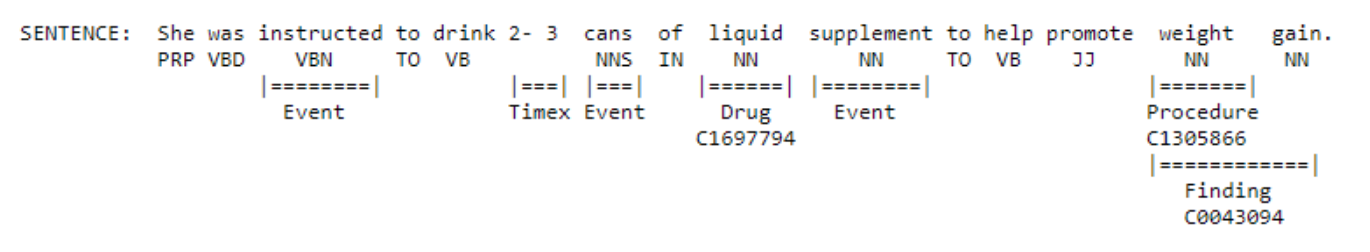}
\caption{An example of cTAKES annotation output with part-of-speech tags and UMLS CUIs for named entities.\footnotemark}
\label{fig:ctakes}
\end{figure*}
\footnotetext{Figure from \url{https://healthnlp.github.io/examples/}.}

cTAKES utilizes a rule-based system for performing ontology mapping, via a UMLS dictionary lookup on the noun phrases inferred by a part-of-speech tagger. Taking raw text as input, the software outputs a set of UMLS concepts identified in the text and their positions, with functionality to map them to other ontologies such as SNOMED and ICD9. It is highly scalable, and can be deployed locally to avoid compromising identifiable patient data. Figure \ref{fig:ctakes} shows an example cTAKES annotation on a clinical record.

\paragraph{Clinical Named-Entity Recognition (NER)}

Recent work has focused on developing tools to replace cTAKES in favor of modern neural architectures such as Bi-LSTM CRFs \cite{boag2018cliner, tao2018effective, xu2018sblc, greenberg2018marginal}, varying in task definition and evaluation. Newer approaches leverage contextualized word embeddings such as ELMo \cite{zhu2018clinical, si2019enhancing}. In contrast, we focus on maximizing the power of existing tools such as cTAKES. This approach is more practical in the near-term, because the adoption of new NER systems in the clinical domain is inhibited by the amount of computational power, data, and gold-label annotations needed to build and train such token-level models, as well as considerations for the effectiveness of domain transfer and a necessity to perform annotations locally in order to protect patient data. Newer models do not provide these capabilities.

\paragraph{NER in Text-based Models}

Prior works use the output of cTAKES as features for disease- and drug-specific tasks, but either concatenate them as shallow features, or substitute them for the text itself (see \citet{wang2017clinical} for a literature review). \citet{weng2017medical} incorporate the output of cTAKES into their input feature vectors for the task of predicting the medical subdomain of clinical notes. However, they use them as shallow features in a non-neural setting, and combine cTAKES annotations with the text representations by concatenating the two into one larger feature vector. In contrast, we propose to learn dense neural concept embedding representations, and integrate the concepts in a learnable fashion to guide the representation learning process, rather than simply concatenating them or using them as a text replacement. We additionally focus on a more challenging task setting.

\citet{boag2017awe} augment a Word2Vec training objective to predict clinical concepts. This work is orthogonal to ours as it is an unsupervised "embedding pretraining" approach rather than an end-to-end supervised model.

\paragraph{Automated Clinical Coding}
The automated clinical coding task is to predict from the raw text of a hospital discharge summary describing a patient encounter all of the possible ICD9 (diagnosis and procedure) codes which a human annotator would assign to the visit. Because these annotators are trained professionals, the ICD codes assigned serve as a natural label set for describing a patient record, and the task can be seen as a proxy for a general patient outcome or treatment prediction task. State-of-the-art methods such as CAML \cite{mullenbach2018explainable} treat each label prediction as a separate task, performing many binary classifications over the many-thousand-dimensional label space. The model is described in more detail in the next section.

\begin{figure*}[ht]
\centering
\includegraphics[width=\linewidth]{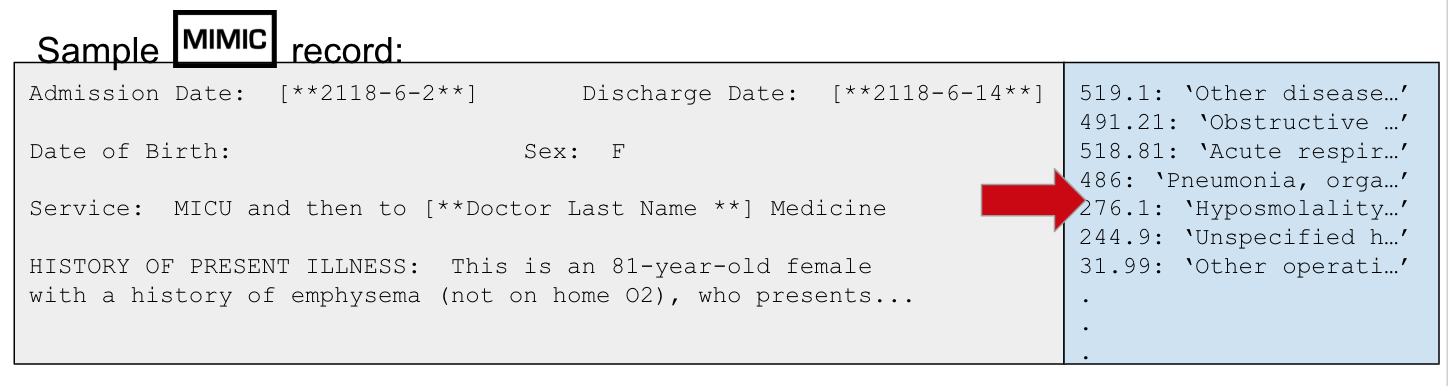}
\caption{An example clinical discharge summary and associated ICD codes.
}
\label{fig:record}
\end{figure*}

The label space is very large (tens of thousands of possible codes) and frequency is long-tailed. \citet{rios2018few} find that CAML performs weakly on rare labels.

\section{Problem Setup}
\label{sec:method}
\paragraph{Task Notation}
\label{ssec:exp:task}
A given discharge summary is represented as a matrix $\bm{X} \in \reals^{d_e\times N}$.\footnote{We use notation for a single instance throughout.} The set of diagnosis and procedure codes assigned to the visit is represented as the one-hot vector $\bm{y} \in \{0,1\}^{L}$. 
The task can be framed as $L = |\set{L}|$ binary classifications: predict $y_{l} \in \{0,1\}$ for code $l$ in labelspace $\set{L}$.

\paragraph{Data} 

We use the publically-available MIMIC-III dataset, a collection of deidentified discharge summaries describing patient stays in the Beth Israel Deaconess Medical Center ICU between 2001 and 2012 \cite{MIMIC_3, mimicdata}. Each discharge summary has been tagged with a set of ICD9 codes. See Figure 3 for an example of a record, and Appendix~\ref{sec:appendix_b} for a description of the dataset and preprocessing.

\label{CE}
\paragraph{Concept Annotation} We run cTAKES on the discharge summaries (described in Appendix \ref{sec:appendix_a}). Results on the extracted concepts are presented in Table \ref{tab:CE}. Note the difference in number of annotations provided by using the SNOMED ontology compared to ICD9.\footnote{Preliminary experiments with sparser ontologies (RXNORM) were not promising, leading us to choose these two ontologies based on their annotation richness (SNOMED) and direct relation to the prediction task (ICD9).}

\begin{table}
\centering
\small
\begin{tabularx}{\columnwidth}{lX}
\toprule
\textbf{ICD9} & \\
Total concepts extracted &     1,005,756     \\
Mean \# extracted concepts per document     &    19.10     \\  
Mean \% words assigned a concept per document & 1.26\% \\ \\
\textbf{SNOMED} &  \\
Total concepts extracted &    28,090,075      \\
Mean \# extracted concepts per document     &     532.76      \\ Mean \% words assigned a concept per document & 35.21\% \\ \\

Mean \# tokens per document & 1513.00 \\
\bottomrule
\end{tabularx}
\caption{Descriptive Statistics on concept extraction for the MIMIC-III corpus.}
\label{tab:CE}
\end{table}


\paragraph{Base model}
We evaluate against CAML~\cite{mullenbach2018explainable}, a state-of-the-art text-based model for the clinical coding task. The model leverages a convolutional neural network (CNN) with per-label attention to predict the combination of codes to assign to a diven discharge summary. Applying convolution over $\bm{X}$ results in a convolved input representation $\bm{H} \in \reals^{d_c\times N}$ (with $d_c < d_e$) in which the column-dimensionality $N$ is preserved. $\bm{H}$ is then used to predict $\bm{y}$, by attentional pooling over the columns.

We include implementation details of all methods, including hyperparameters and training, in Appendix \ref{sec:appendix_b}.


\section{Approach 1: Augmentation Model}
\label{sec:aug}
One limitation of learning-based models is their tendency to lose uncommon words to "UNK" tokens, or to suffer from poor representation learning for them. We hypothesize that rare words are important for predicting rare labels, and that text-based models may be improved by augmenting word embeddings with concept embeddings as a means to strengthen representations of rare or unseen words. We additionally hypothesize that linking multiple words to a shared concept via cTAKES annotation will reduce textual noise by grouping word variants to a shared representation in a smaller and more frequently updated parameter space.


\subsection{Method}

Given a discharge summary containing words $w_1, w_2, ..., w_N \in \set{W}^*$ and an embedding function $\gamma: \set{W} \rightarrow \reals^{d_e}$, we construct input matrix $\bm{X} = [\bm{x}_1^T, \bm{x}_2^T, ..., \bm{x}_N^T] \in \reals^{d_e\times N}$ as column-stacked word embeddings, where $\bm{x}_{n} = \gamma (w_{n})$.

We additionally assume a code embedding function $\phi: \set{C} \to \reals^{d_e}$ and a set of annotated codes for a given document $c_1, c_2, \ldots, c_N \in \set{C^*}$, where $\set{C}$ is the full codeset for the ontology used to annotate the document, and $c_n$ is the code annotated for word token $w_n$, if one exists (else $c_n = \varnothing$, by abuse of notation). 
 We construct a representation for each document, $\bm{D}$, of the same dimensionality as  $\bm{X}$, by learning one representation leveraging both the concept and word embedding at each position:
 

For token $n$,
\begin{align}
\bm{d}_n = &{} \beta_{w_n,c_n} \bm{\phi}(c_n) + (1 - \beta_{w_n, c_n}) \bm{x}_n,
\end{align}
$\beta_{w_n,c_n} \in [0,1]$ is a learned parameter specific to each observed word+concept pair, including UNK tokens.\footnote{We experimented with models in which this gate was computed element-wise and shared by all word+concept pairs (e.g. by passing $\bm{x}_n$ and $\phi(c_n)$ through a linear layer or simple multi-layer perceptron to compute $\bm{d}_n$), but this did not improve performance.} Intuitively, if there is a concept associated with index $n$, a concept embedding $\bm{\phi}(c_n)$ is generated and a \emph{linear combination} of the word and concept embedding is learned, using a learned parameter specific to that word+concept pair.\footnote{A single token may have multiple concept annotations associated with it. We experiment with an attention mechanism for this case (see Appendix \ref{ssec:method:attn}), but find a heuristic of arbitrarily selecting the first concept assigned to each word performs just as well.} We fix $\beta_{w_n, c_n = \varnothing} = 0$, which reverts to the word embedding when there is no concept assigned. 

We additionally propose a simpler version of this method, \emph{full replace}, in which word embeddings are completely replaced with concept embeddings if they exist (i.e. $\beta_{w_n, c_n} = 1, \, \forall w_n, c_n \neq \varnothing$).
In this formulation, if a concept spans multiple words, all of those words are represented by the same vector.
Conversely, the CAML baseline corresponds to a model in which $\beta_{w_n,c_n} = 0 , \, \forall w_n, c_n$.  





\subsection{Evaluation Setup}
\label{ssec:experiment}

\paragraph{Metrics}

In addition to the metrics reported in prior work, we report average precision score (AP), which is a preferred metric to AUC for imbalanced classes \cite{saito2015precision, davis2006relationship}. We report both macro- and micro- metrics, with the former being more favorable toward rare labels by weighting all classes equally.
We additionally focus on the precision-at-$k$ (P@$k$) metric, representing the fraction of the $k$ highest-scored predicted labels that are present in the ground truth. Both macro-metrics and P@$k$ are useful in a computer-assisted coding use-case, where the desired outcome is to correctly identify needle-in-the-haystack labels as opposed to more frequent ones, and to accurately suggest a small subset of codes with the highest confidence as annotation suggestions \cite{mullenbach2018explainable}.

\paragraph{Baselines}

Along with CAML, we evaluate on a \emph{raw codes} baseline where the ICD9 annotations generated by cTAKES $c_1, c_2, \ldots, c_N$ are used directly as the document-level predictions. Formally, $\hat y_{c_n} = 1 \text{ when } c_n \in \set{L} \text{ and } c_n \neq \varnothing$, for all $n$ in integers $1$ to $N$.
\subsection{Results}
\label{ssec:aug_results}
\begin{table*}[ht]
\centering
\resizebox{\textwidth}{!}{\begin{tabular}
{l|ll|ll|ll|ll|ll}
\toprule
& \multicolumn{2}{c}{\textbf{AUC}} & \multicolumn{2}{c}{\textbf{AP}} & \multicolumn{2}{c}{\textbf{F1}} & \multicolumn{2}{c}{\textbf{R@k}} & \multicolumn{2}{c}{\textbf{P@k}}\\
 \textbf{Model} & Macro & Micro & Macro & Micro & Macro & Micro & 8 & 15 & 8 & 15 \\ \midrule
Baseline \cite{mullenbach2018explainable} & 0.8892 & 0.9846 & \textbf{0.2492} & 0.5426 & \textbf{0.0796} & \textbf{0.5421} & \textbf{0.3731} & \textbf{0.5251} & 0.7120 & 0.5616 \\
Baseline (raw codes) & n/a$^*$ & n/a$^*$ & n/a$^*$ & n/a$^*$ & 0.0189 & 0.0877 & 0.0534$^*$ & 0.0640$^*$ & 0.1132$^*$ & 0.0747$^*$\\ \hline
\textbf{Augmentation with ICD9} \\
full replace & 0.8846 & 0.9838 & 0.2242 & 0.5329 & 0.0691 & 0.5363 & 0.3688 & 0.5189 & 0.7048 & 0.5564 \\
linear combination & \textbf{0.8914} & \textbf{0.9849} & 0.2467 & \textbf{0.5427} & 0.0763 & 0.5419 & \textbf{0.3732} & \textbf{0.5267} & \textbf{0.7121} & \textbf{0.5634} \\ \hline
\textbf{Augmentation with SNOMED}\\
full replace & 0.8744 & 0.9830 & 0.2221 & 0.5271 & 0.0724 & 0.5326 & 0.3675 & 0.5177 & 0.7022 & 0.5547\\
linear combination & 0.8781 & 0.9835 & 0.2238 & 0.533 & 0.0692 & 0.5357 & 0.3687 & 0.5194 & 0.7042 & 0.5563\\
\bottomrule
\end{tabular}}
\caption{Test set results using the augmentation methods.}
\label{tab:aug_results}
\end{table*}

\blfootnote{*These metrics were computed by randomly selecting $k$ elements from those predicted, since there are no sorted probabilities associated with this baseline. For the same reasons we cannot report AUC or AP metrics.}

We present results on the test set in \autoref{tab:aug_results}. Overall, the concept-augmented models are indistinguishable from the baseline, and there is no significant difference between annotation type or recombination method, although the linear combination method with ICD9 annotations is the best performing and rivals the baseline. 

Following the negative results for our initial attempt to augment word embeddings with concept embeddings, we tried two alternative strategies:
\begin{itemize}
\item We concatenated the ICD9 annotations with two other ontologies: RXNORM and SNOMED. While this led to greater coverage over the text (with slightly more than one third of the tokens in the text receiving corresponding concept annotations), it did not improve downstream performance.
\item 
Prior work has demonstrated that leveraging clinical ontological structure can allow models to learn more effective code embeddings in fully structured data models \cite{singh2014leveraging, choi2017gram}. We applied the methodology of \citet{choi2017gram} on both the ICD9 and SNOMED annotations, but this did not improve performance. For more details, see Appendix \ref{ssec:gram}. 
\end{itemize}

\subsection{Error Analysis}
\label{ssec:error_analysis}

\begin{figure}
\centering
\includegraphics[width=0.9\linewidth]{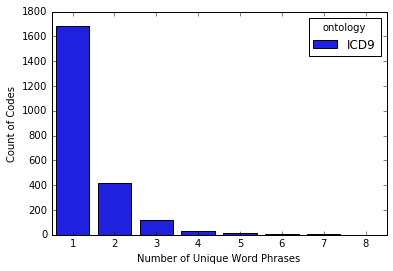}
\caption{A histogram showing the distribution of ICD9 concepts in $\set{C}$ grouped according to the number of unique word phrases in the MIMIC-III corpus associated with each. We observe the same trend when plotting SNOMED annotations.}
\label{fig:word-to-concept}
\end{figure}

Error analysis of the word-to-concept mapping produced by cTAKES exposes limitations of our initial hypothesis that cTAKES mitigates word-level variation by assigning multiple distinct word phrases to shared concepts. Figure \ref{fig:word-to-concept} demonstrates that the vast majority of the ICD9 concepts in the corpus are assigned to only one distinct word phrase, and the same results are observed for SNOMED concepts. This may explain the virtually indistinguishable performance of the augmentation models from the baseline, because randomly-initialized word and concept embeddings which are observed in strictly identical contexts should theoretically converge to the same representation.\footnote{Simulations of the augmentation method under a contrived setting with more concept annotations per note as well as more unique word phrases mapping to a single concept demonstrate solid performance increases over the baseline. This provides supporting evidence that the findings presented in this section may be the cause of the negative result rather than our proposed architecture.} 

The raw codes baseline performs poorly, which aligns with the observation that cTAKES codes assigned to a discharge summary often do not have appropriate or proportional levels of specificity (for example, the top-level ICD9 code '428 Heart Failure' may be assigned by cTAKES, but the gold-label code is '428.21 Acute Systolic Heart Failure'). This may also contribute to the negative result of the proposed model.

Figure \ref{fig:f1} 
(included in the Appendix) illustrates prediction performance as a function of code frequency in the training set, showing that the proposed model does not improve upon the baseline for rare or semi-rare codes.\footnote{We use the following grouping criteria: rare codes have 50 or fewer occurrences in the training data, semi-rare have between 50 and 1000, and common have more than 1000.}

\subsection{Ablations}
\label{sec:ablations}

We separate and analyze the two distinct components of cTAKES' annotation ability for further analysis: 1) how well cTAKES recognizes the location of concepts in the text (\emph{NER}), and 2) how accurately cTAKES maps the recognized positions to the correct clinical concepts (\emph{ontology mapping}). Annotation sparsity (NER) and/or cTAKES mapping error may lend the raw text on its own equally useful, as observed in Table \ref{tab:aug_results}. We investigate these hypotheses here. We evaluate performance of ablations relative to \mname and baseline to determine whether each component individually adds value. The ablations are:

\begin{enumerate}
    \item \emph{Dummy Concepts} We replace all word embeddings annotated by cTAKES with 0-vectors, and only use remaining embeddings for prediction. If this alternative shows similar performance to the baseline, then we conclude that the positions in the text annotated by cTAKES (NER) are not valuable for prediction performance.
    \item \emph{Concepts Only} We test the complement by replacing all word embeddings \textit{not} annotated by cTAKES with a 0-vector. In contrast to Dummy Concepts, strong performance of this approach relative to the baseline will allow us to conclude that the positions in the text annotated by cTAKES are valuable for prediction performance.
    \item \emph{Concepts Only, Concept Embeddings} We replace all word embeddings not annotated by cTAKES with a 0-vector, and then replace all remaining word embeddings with their concept embedding. If this model performs better than Concepts Only, it will demonstrate the strength of cTAKES' ontology mapping component. 
\end{enumerate}

Note that Dummy Concepts and Concepts Only are the decomposition of the baseline CAML. Similarly, Dummy Concepts and Concepts Only, Concept Embeddings are the decomposition of the full-replace augmentation model presented in Section \ref{sec:aug}.

\begin{table*}[ht]
\centering
\resizebox{\textwidth}{!}{\begin{tabular}{l|ll|ll|ll|ll|ll|ll}
\toprule
 & \multicolumn{2}{c}{\textbf{Token Representation}} & \multicolumn{2}{c}{\textbf{AUC}} & \multicolumn{2}{c}{\textbf{AP}} & \multicolumn{2}{c}{\textbf{F1}} & \multicolumn{2}{c}{\textbf{R@k}} & \multicolumn{2}{c}{\textbf{P@k}}\\
\textbf{Model} & Concept Match & No Match & Macro & Micro & Macro & Micro & Macro & Micro & 8 & 15 & 8 & 15 \\ \midrule
Baseline \cite{mullenbach2018explainable} & Word & Word & \textbf{0.8892} & \textbf{0.9846} & \textbf{0.2492} & \textbf{0.5426} & \textbf{0.0796} & \textbf{0.5421} & \textbf{0.3731} & \textbf{0.5251} & \textbf{0.7120} & \textbf{0.5616} \\
Dummy Concepts & $\bm{0}$ & Word & 0.8876 & 0.9839 & 0.2119 & 0.5236 & 0.0732 & 0.5261 & 0.3634 & 0.5141 & 0.6943 & 0.5506 \\
Concepts Only & Word & $\bm{0}$ & 0.7549 & 0.9626 & 0.0538 & 0.2487 & 0.0080 & 0.1961 & 0.2063 & 0.2880 & 0.4196 & 0.3197 \\
Concepts Only, Concept Embeddings & Concept & $\bm{0}$ & 0.7534 & 0.9620 & 0.0552 & 0.2464 & 0.0086 & 0.1972 & 0.2058 & 0.2855 & 0.4200 & 0.3166\\
Augmentation Model (full replace) & Concept & Word & 0.8846 & 0.9838 & 0.2242 & 0.5329 & 0.0691 & 0.5363 & 0.3688 & 0.5189 & 0.7048 & 0.5564 \\
\bottomrule
\end{tabular}}
\caption{Test set results of ablation experiments on the MIMIC-III dataset, using ICD9 concept annotations.}
\label{tab:ICD_abl}
\end{table*}
\begin{table*}[ht]
\centering
\resizebox{\textwidth}{!}{\begin{tabular}{l|ll|ll|ll|ll|ll|ll}
\toprule
 & \multicolumn{2}{c}{\textbf{Token Representation}} & \multicolumn{2}{c}{\textbf{AUC}} & \multicolumn{2}{c}{\textbf{AP}} & \multicolumn{2}{c}{\textbf{F1}} & \multicolumn{2}{c}{\textbf{R@k}} & \multicolumn{2}{c}{\textbf{P@k}}\\
\textbf{Model} & Concept Match & No Match & Macro & Micro & Macro & Micro & Macro & Micro & 8 & 15 & 8 & 15 \\ \midrule
Baseline \cite{mullenbach2018explainable} & Word & Word &  \textbf{0.8892} & \textbf{0.9846} & \textbf{0.2492} & \textbf{0.5426} & \textbf{0.0796} & \textbf{0.5421} & \textbf{0.3731} & \textbf{0.5251} & \textbf{0.7120} & \textbf{0.5616} \\
Dummy Concepts & $\bm{0}$ & Word & 0.8472 & 0.9780 & 0.1461 & 0.4375 & 0.0413 & 0.4426 & 0.3202 & 0.4439 & 0.6234 & 0.4804\\
Concepts Only & Word & $\bm{0}$ & 0.8736 & 0.9817 & 0.2059 & 0.4518 & 0.0515 & 0.4295 & 0.3278 & 0.4583 & 0.6300 & 0.4903\\
Concepts Only, Concept Embeddings & Concept & $\bm{0}$ & 0.8739 & 0.9813 & 0.2019 & 0.4451 & 0.0519 & 0.4258 & 0.3247 & 0.4538 & 0.6254 & 0.4851\\
Augmentation Model (full replace) & Concept & Word & 0.8744 & 0.9830 & 0.2221 & 0.5271 & 0.0724 & 0.5326 & 0.3675 & 0.5177 & 0.7022 & 0.5547\\
\bottomrule
\end{tabular}}
\caption{Test set results of ablation experiments on the MIMIC-III dataset, using SNOMED concept annotations.}
\label{tab:SNOMED_abl}
\end{table*}
\paragraph{Results}

Results are presented in Tables \ref{tab:ICD_abl} and \ref{tab:SNOMED_abl}. Results are consistent with previous experiments in that augmentation with concept annotations does not improve performance. For both ontologies, neither the Dummy Concepts nor the Concepts Only models outperform the full-text models (in which both token representations are used). However, there are some interesting findings. Using SNOMED annotations, performance of the Concepts Only model is significantly higher than Dummy Concepts and very close to full-text model performance. This finding is strengthened by considering the concept coverage discussed in Table \ref{tab:CE}: the Concepts Only model achieves comparable performance receiving only about 35\% (1\% in the ICD9 setting) of the input tokens which the full-text baseline receives, and the Dummy Concepts Model receives about 65\% (99\% in the ICD9 setting). Thus, a significant proportion of downstream prediction performance can be attributed a small portion of the text which is recognized by cTAKES in both the SNOMED and ICD9 settings, indicating the strength of cTAKES' NER component. 

\section{Approach 2: Multi-task Learning}
\label{sec:mtl}


We present an alternative application of cTAKES as a form of distant supervision. Our approach is inspired by recent successes in multi-task learning for NLP which demonstrate that cheaply-obtained labels framed as an auxiliary task can improve performance on downstream tasks \cite{swayamdipta2018syntactic, ruder2017overview, zhang2016stack}. We propose to predict clinical information extraction system annotations as an auxiliary task, and share lower-level representations with the clinical coding task through a jointly-trained model architecture. We hypothesize that domain-knowledge embedded in cTAKES will guide the shared layers of the model architecture towards a more optimal representation for the clinical coding task.

We formulate the auxiliary task as follows: given each word-embedding or word-embedding span in the input which cTAKES has assigned a code, can the model predict the code assigned to it by cTAKES?  

\subsection{Method}

We annotate the set of non-null ground-truth codes output by cTAKES for document $i$ in the training data as $\{(a_{i,1},c_{i,1}), (a_{i,2}, c_{i,2}),\ldots, (a_{i,M}, c_{i,M})\}$, where each anchor $a_{i,m}$ indicates the span of tokens in the text for which concept $c_{i,m}$ is annotated, and $c_{i,m} \neq \varnothing$. 

The loss term of the model is augmented to include the multi-class cross-entropy of predicting the correct code for all annotated spans in the training batch:
\begin{align*}
    \mathcal{L}  = \sum_{i=1}^I &{} BCE(\bm{y}_i, \hat{\bm{y}_i})\\
                  &{} + \lambda\frac{\sum_{i=1}^{I}\sum_{m=1}^{M_i}-\log p(c_{i,m}\mid a_{i,m})}{\sum_{i=1}^{I} M_i}
\end{align*}
where $BCE(\bm{y}_i, \hat{\bm{y}_i})$ is the standard (binary cross-entropy) loss from the baseline for the clinical coding task,  $p(c_{i,m}\mid a_{i,m})$ is the probability assigned by the auxiliary model to the true cTAKES-annotated concept given word span $a_{i,m}$ as input, $\lambda$ is the hyperparameter to tradeoff between the two objectives, and $I$ is the number of instances in the batch.

\begin{figure}[ht]
\centering
\includegraphics[width=\linewidth]{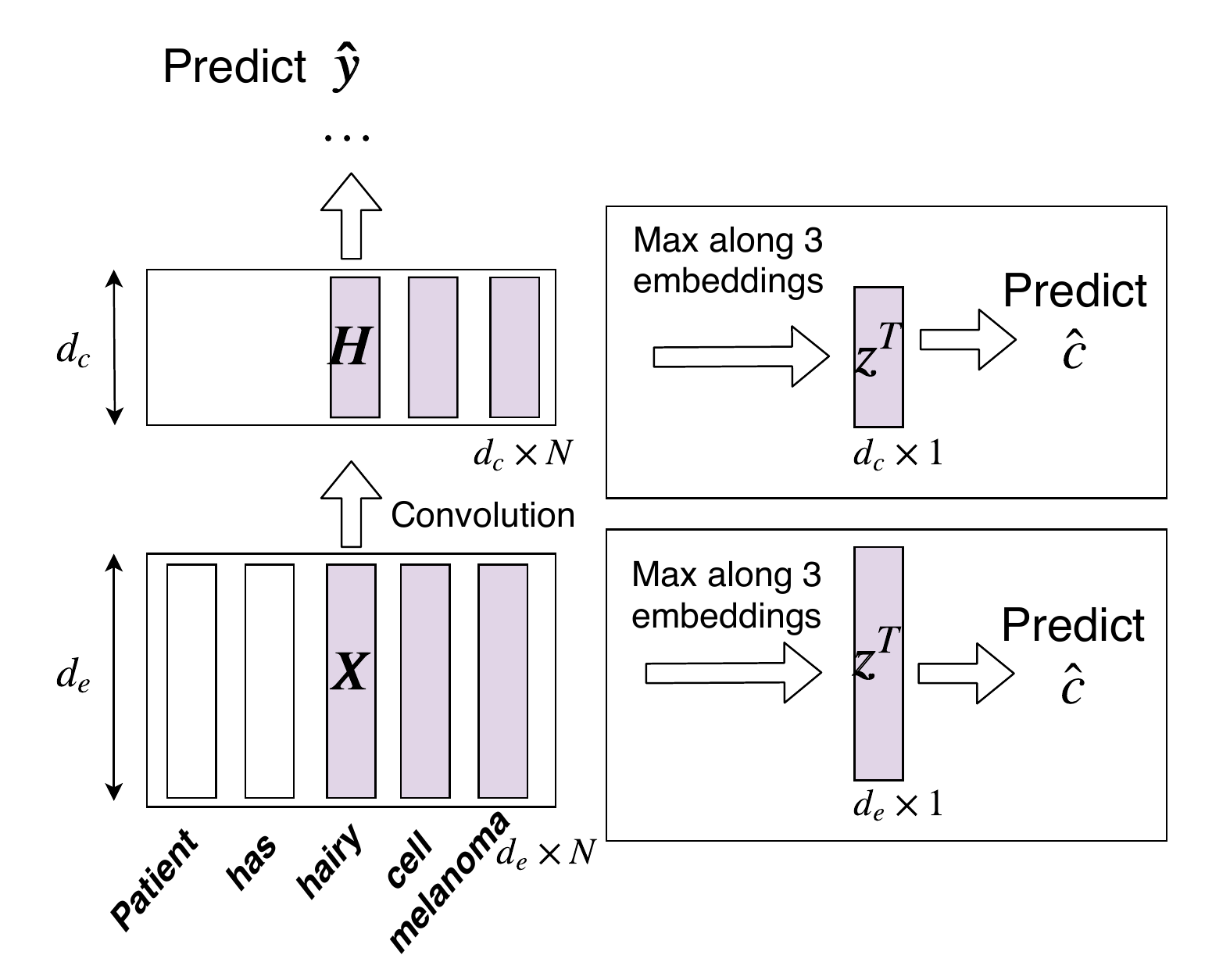}
\caption{The proposed architecture (for prediction on a single document, $i$, and auxiliary supervision on a single annotation, $m$). The bottom box illustrates the pre-convolution model, and the top box post-convolution. The architecture on the left is the baseline.}
\label{fig:mtl_arch}
\end{figure}
\begin{table*}[ht]
\centering
\resizebox{\textwidth}{!}{\begin{tabular}{l|l|ll|ll|ll|ll|ll}
\toprule
 &  & \multicolumn{2}{c}{\textbf{AUC}} & \multicolumn{2}{c}{\textbf{AP}} & \multicolumn{2}{c}{\textbf{F1}} & \multicolumn{2}{c}{\textbf{R@k}} & \multicolumn{2}{c}{\textbf{P@k}} \\
\textbf{Shared Features} & \textbf{Auxiliary Model} & Macro & Micro & Macro & Micro & Macro & Micro & 8 & 15 & 8 & 15 \\
\midrule
Baseline \cite{mullenbach2018explainable} & n/a & \textbf{0.8892} & \textbf{0.9846} & \textbf{0.2492} & \textbf{0.5426} & \textbf{0.0796} & \textbf{0.5421} & \textbf{0.3731} & 0.5251 & \textbf{0.7120} & 0.5616 \\
Pre-convolution & MLP &  0.8874 & 0.9839 & 0.2365 & 0.5390 & 0.0734 & 0.5376 & 0.3724 & 0.5235 & 0.7102 & 0.5597\\
Pre-convolution & Linear Layer & 0.8834 & 0.9838 & 0.2398 & 0.5412 & 0.0766 & 0.5414 & \textbf{0.3731} & \textbf{0.5265} & 0.7113 & \textbf{0.5633}\\
Post-convolution & MLP & 0.7252 & 0.9619 & 0.0578 & 0.3002 & 0.0159 & 0.2966 & 0.2449 & 0.3417 & 0.4879 & 0.3748\\
Post-convolution & Linear Layer & 0.7562 & 0.9655 & 0.0606 & 0.3035 & 0.0123 & 0.2934 & 0.2461 & 0.3392 & 0.4900 & 0.3700\\
\bottomrule
\end{tabular}}
\caption{Test set performance on the ICD9 coding task for $\lambda=1$ and using ICD9 annotations.}
\label{tab:aux_main}
\end{table*}

\begin{table}[ht]
\centering
\resizebox{\columnwidth}{!}{\begin{tabular}{l|l|l|l}
\toprule
 &  & \multicolumn{2}{c}{\textbf{Tagging Accuracy}}\\ 
\textbf{Shared Features} & \textbf{Auxiliary Model} & \textbf{After one epoch} &  \textbf{After last epoch}\\ 
\midrule
Pre-convolution & MLP & 0.9343 & 0.9398 \\
Pre-convolution & Linear Layer & 0.8940 & \textbf{0.9400} \\
Post-convolution & MLP & 0.9102 & 0.9335 \\
Post-convolution & Linear Layer & 0.7524 & 0.9341 \\
\bottomrule
\end{tabular}}
\caption{Dev set performance on the auxiliary task for $\lambda=1$ and using ICD9 annotations. Relatively high task performance is achieved even after one epoch with a simple model.}
\label{tab:aux_aux}
\end{table}

Because we use the auxiliary task as a ``scaffold'' \cite{swayamdipta2018syntactic} for transferring domain knowledge encoded in cTAKES' rules into the learned representations for the clinical coding task, we must only run cTAKES and compute a forward pass through the auxiliary module at training time. At test-time, we evaluate only on the clinical coding task, so the time complexity of model inference remains the same as the baseline, an advantage of this architecture.

We model $p(c_{i,m}\mid a_{i,m})$ via a multi-layer perceptron with a Softmax output layer to obtain a distribution over the codeset, $\set{C}$. We additionally experiment with a linear layer variant to combat overfitting on the auxiliary task by reducing the capacity of this module. The input to this module is a single vector, $\bm{z}_{i,m} \in \reals^{d_e}$, constructed by selecting the maximum value over $s$ word embeddings for each dimension, where $s$ is the length of the input span.\footnote{While this is simple representation, we find that multi-word concept annotations are rather rare, in which case $\bm{z}_{i,m}$ is equivalent to $\bm{x}_{i,m}$.} To facilitate information transfer between the clinical coding and auxiliary task, we experiment with tying both the randomly-initialized embedding layer, $\bm{X}$, and a higher-level layer of the network (e.g. the outputs of the document-level convolution layer $\bm{H}$ described in Section \ref{ssec:exp:task}). See Figure \ref{fig:mtl_arch} for the model architecture.

\subsection{Experiment and Results}

Results are presented in \autoref{tab:aux_main} and \autoref{tab:aux_aux} for ICD9 annotations.
Overall, 
the cTAKES span-prediction task does more to hurt than help performance on the main task.
Tying the model weights at a higher layer (post-convolution as opposed to pre-convolution) results in worse performance, even though the model fits the auxiliary task well. This indicates either that the model may not have enough capacity to adequately fit both tasks, or that the cTAKES prediction task as formulated may actually misguide the clinical coding task slightly in parameter search space.\footnote{We found similar results using SNOMED annotations.}

 We additionally remark that increasing the weight of the auxiliary task generally lowers performance on the clinical coding task, and tuning $\lambda$ on the dev set does not result in more optimal performance (we include results with $\lambda=1$ here; see \autoref{tab:aux_addtl} in the Appendix). Notably, for even very small values of $\lambda$, we achieve very high validation accuracy on the auxiliary task. This performance does not change with larger weightings, indicating that the auxiliary task may not be difficult enough to result in effective knowledge transfer.\footnote{While the models in Sections \ref{sec:aug} did not introduce new hyperparameters to the baseline architecture, hyperparameters for this architecture were selected by human intuition. Room for future work includes more extensive tuning (see Table \ref{tab:model-details} in Appendix \ref{sec:appendix_b}).} 



\section{Conclusion}
\label{sec:conclusion}
\label{sec:discussion}
Integrating existing clinical information extraction tools with deep learning models is an important direction for bridging the gap between rule-based and learning-based methods. We have provided an analysis of the quality of the widely-used clinical concept annotator cTAKES when integrated into a state-of-the-art text-based prediction model. In two settings, we have shown that cTAKES does not improve performance over raw text alone on the clinical coding task.
We additionally demonstrate through error analysis and ablation studies that the amount of word variation captured and the differentiation between the named-entity recognition and ontology-mapping tasks may affect cTAKES' effectiveness.

While automated coding is one application area, the models presented here could easily be extended to other downstream prediction tasks such as patient diagnosis and treatment outcome prediction. Future work will include evaluating newly-developed clinical NER tools with similar functionalities to cTAKES in our framework, which can potentially serve as a means to evaluate the effectiveness of newer systems vis-\`a-vis cTAKES.







\paragraph{Acknowledgments}
We thank the Research, Development, and Dissemination group at Sutter Health as well as members of the Georgia Tech Computational Linguistics Lab  for helpful discussions. Figure 3 was provided by James Mullenbach.

\bibliography{acl2019}
\bibliographystyle{acl_natbib}

\clearpage
\appendix

\section{Experimental Details}
\label{sec:appendix_b}
\paragraph{Data}
Following \citet{mullenbach2018explainable}, we use the same train/test/validation splits for the MIMIC-III dataset, and concatenate all supplemental text for a patient discharge summary into one record. We use the authors' provided data processing pipeline\footnote{\url{https://github.com/jamesmullenbach/caml-mimic/blob/master/notebooks/dataproc_mimic_III.ipynb}} to preprocess the corpus. The vocubulary includes all words occurring in at least 3 training documents. See Table \ref{tab:stats} for descriptive statistics of the dataset.

We construct a concept vocabulary for embedding initialization following the same specification as the word vocabulary: any concept which does not occur in at least 3 training documents is replaced with an UNK token. Details on the size of the vocabulary can be found in Table \ref{tab:model-details}. 

\begin{table}[ht]
\centering
{\begin{tabularx}{\columnwidth}{ll}
\toprule
\# training documents                &  47,723   \\
\# test documents                &  3,372    \\
\# dev documents                &  1,631   \\
\midrule
Mean \# tokens per document & 1,513.0  \\
Mean \# labels per document & 16.09 \\
Total \# labels ($L$)            & 8,921  \\
\bottomrule
\end{tabularx}}
\caption{Dataset Descriptive Statistics.}
\label{tab:stats}
\end{table}

\paragraph{Training}
We train with the same specifications as \citet{mullenbach2018explainable} unless otherwise specified, with dropout performed after concept augmentation for the models in Sections \ref{sec:aug}, and early stopping with a patience of 10 epochs on the precision at 8 metric, for a maximum of 200 epochs (note that in the multi-task learning models the stopping criterion is only a function of performance on the clinical coding task). Unlike previous work, we reduce the batch size to 12 in order to allow each batch to fit on a single GPU, and we do not use pretrained embeddings as we find this improves performance. All models are trained on a single NVIDIA Titan X GPU with 12,189 MiB of RAM. 

We port the optimal hyperparameters reported in \citet{mullenbach2018explainable} to our experiments. With more extensive hyperparameter tuning, we may expect to see a potential increase in the performance of our models over the baseline. See Table \ref{tab:model-details} for hyperparameters and other details specific to our proposed model architectures. All neural models are implemented using PyTorch\footnote{\url{https://github.com/pytorch/pytorch}}, and built on the open-source implementation of CAML.\footnote{\url{https://github.com/jamesmullenbach/caml-mimic}}

\begin{table}[ht]
\centering
\small
\begin{tabularx}{\columnwidth}{Xr}
\toprule
Parameter & Value \\ \midrule
Vocabulary Size & 51,917 \\
SNOMED Concept Vocabulary ($\set{C}$) Size & 20,775 \\
ICD9 Concept Vocabulary ($\set{C}$) Size & 1,529 \\ \hline
Embedding Size ($d_e$) & 100 \\
Post-convolution Embedding Size ($d_c$) & 50\\
Dropout Probability & 0.2 \\
Learning Rate & 0.0001\\ \hline
Attention Mechanism Hidden State Size & 20 \\
Attention Mechanism Activation Function & ReLU \\
Auxiliary hidden layer size & 700 \\
Auxiliary activation function & ReLU \\
\bottomrule
\end{tabularx}
\caption{Model details.}
\label{tab:model-details}
\end{table}



\begin{table*}[ht]
\centering
\resizebox{\textwidth}{!}{\begin{tabular}{l|ll|ll|ll|ll|ll|ll}
\toprule
$\lambda$ & \multicolumn{2}{c}{\textbf{AUC}} & \multicolumn{2}{c}{\textbf{AP}} & \multicolumn{2}{c}{\textbf{F1}} & \multicolumn{2}{c}{\textbf{R@k}} & \multicolumn{2}{c}{\textbf{P@k}} & \multicolumn{2}{c}{\textbf{Auxiliary Tagging Accuracy}}\\
& Macro & Micro & Macro & Micro & Macro & Micro & 8 & 15 & 8 & 15 & After one epoch & After last epoch \\
\midrule
0.001 & \textbf{0.9002} & \textbf{0.9848} & 0.3129 & 0.5470 & \textbf{0.0704} & \textbf{0.5511} & 0.3902 & 0.5447 & 0.7164 & 0.5631 & 0.8888 & 0.9398 \\
0.01 & 0.8954 & 0.9842 & 0.2885 & 0.5352 & 0.0636 & 0.5425 & 0.3843 & 0.5328 & 0.7088 & 0.5528 & 0.8938 & \textbf{0.9401} \\
0.1 & 0.9000 & 0.9846 & \textbf{0.3145} & 0.5465 & 0.0689 & 0.5471 & \textbf{0.3909} & 0.5426 & \textbf{0.7183} & 0.5617 & 0.8940 & 0.9400 \\
0.5 & 0.8934 & 0.9840 & 0.2892 & 0.5362 & 0.0624 & 0.5386 & 0.3844 & 0.5361 & 0.7089 & 0.5546 & \textbf{0.8941} & 0.9400 \\
1 & 0.8975 & 0.9840 & 0.3087 & 0.5460 & 0.0668 & 0.5477 & 0.3886 & 0.5439 & 0.7169 & 0.5624 & 0.8940 & 0.9400 \\
10 & 0.8979 & 0.9842 & 0.3122 & \textbf{0.5484} & 0.0678 & 0.5474 & 0.3908 & \textbf{0.5457} & 0.7182 & \textbf{0.5644} & 0.8940 & 0.9400 \\
50 & 0.8939 & 0.9837 & 0.2982 & 0.5410 & 0.0638 & 0.5427 & 0.3855 & 0.5391 & 0.7111 & 0.5592 & 0.8940 & \textbf{0.9401} \\
100 & 0.8913 & 0.9835 & 0.2943 & 0.5383 & 0.0632 & 0.5407 & 0.3849 & 0.5374 & 0.7096 & 0.5577 & 0.8940 & \textbf{0.9401} \\
1000 & 0.8851 & 0.9827 & 0.2750 & 0.5260 & 0.0564 & 0.5309 & 0.3803 & 0.5290 & 0.7016 & 0.5491 & 0.8940 & \textbf{0.9401} \\
\bottomrule
\end{tabular}}
\caption{The effect of tuning $\lambda$ on dev set performance on the ICD9 coding task, for the pre-convolution model with a linear auxiliary layer and ICD9 annotations. We select $\lambda=1$ for reporting test results; there isn't a clear value which produces strictly better performance.}
\label{tab:aux_addtl}
\end{table*}

\section{Concept Extraction}
\label{sec:appendix_a}
We build a custom dictionary from the UMLS Metathesaurus that includes mappings from UMLS CUIs to SNOMED-CT and ICD9-CM concepts. We run the cTAKES annotator in advance of training for all 3 dataset splits using the resulting dictionary, allowing us to obtain annotations for each note in the dataset, and the positions of the annotations in the raw text. Note that for the multi-task learning experiments (Section \ref{sec:mtl}), we only require annotations for training data.
Annotating the MIMIC-III datafiles using these specifications takes between 4 and 5 hours for 3,000 discharge summaries on a single CPU, and can be parallelized for efficiency. 

\section{Attention for Overlapping Concepts}
\label{sec:appendix_att}
\label{ssec:method:attn}

We implement an attention mechanism \cite{bahdanau2014neural} to compute a single concept embedding $\phi(\set{C}_n) \in \reals^{d_e}$ when $\set{C}_n = \{c_1,c_2,\ldots,c_J\}$ represents a set of concepts annotated at position $n$ instead of a single concept. Intuitively, we want to more heavily weight those concepts in the set which have the most similarity to the surrounding text. We define a context vector for position $n$ as:
\begin{align*}
\bm{v}_{n} = [\bm{x}_{n-2},\bm{x}_{n-1},\bm{x}_{n+1},\bm{x}_{n+2}] \in \reals^{4d_e}
\end{align*}

The context is defined as the concatenated word embeddings surrounding position $n$. We use a context size of $n+/-2$, where $2$ is a hyperparameter. We choose to use a smaller value for computational efficiency.

We concatenate the word-context vector and each concept embedding $c_j$ in $\set{C}_n$ as $[\bm{v}_n, \phi(c_j)] \in \reals^{5d_e}$, and pass it through a multi-layer perceptron to compute a similarity score: $f:\reals^{5d_e} \rightarrow \reals^1$. An attention score for each $c_j$ is computed as:
\begin{align*}
\alpha_{j} = \frac{exp(f(\bm{v}_n,\phi(c_j))}{\sum_{k=1}^{J} exp(f(\bm{v}_n,\phi(c_k))}
\end{align*}

This represents the relevance of the concept to the surrounding word-context, normalized by the other concepts in the set. A final concept embedding $\phi(\set{C}_n) \in \reals^{d_e}$ is computed as a linear combination of the concept vectors, weighted by their attention scores:
\begin{align*}
\phi(\set{C}_n) = \sum_{j=1}^{J} \alpha_{j} \cdot \phi(c_j)
\end{align*}

\section{Leveraging Ontological Graph Structure}
\label{sec:appendix_c}

\label{ssec:gram}

Following the methodology of \citet{choi2017gram}, we experiment with learning higher-quality concept representations using the hierarchical structure of the ICD9 ontology. We replace concept embedding $\phi(c_n)$ with a learned linear combination of itself and its parent concepts' embeddings (see \autoref{fig:icd}). For child concepts which are observed infrequently or have poor representations, prior work has shown that a trained model will learn to weight the parent embeddings more heavily in the linear combination. Because the parent concepts represent more general concepts, they have most often been observed more frequently in the training set and have stronger representations. This also allows for learned representations which capture relationships between concepts. We refer the reader to \citet{choi2017gram} for details. 


\begin{figure}[ht]
\centering
\includegraphics[width=0.48\textwidth]{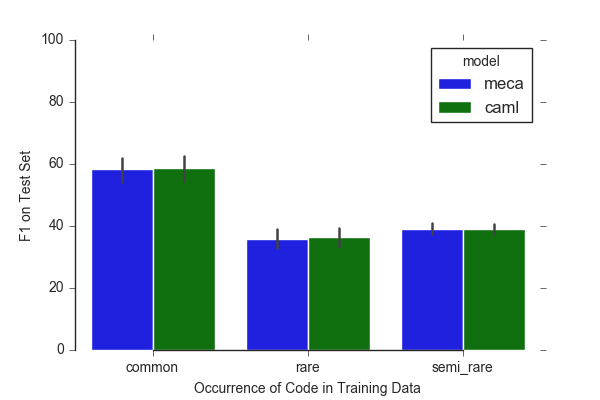}
\caption{F1 on Test Data based on Frequency of Codes in Training Data, where the metric is defined ('meca' indicates the \emph{linear combination} ICD9 augmentation model).}
\label{fig:f1}
\end{figure}



\end{document}